# Objective Bicycle Occlusion Level Classification using a Deformable Parts-Based Model

Angelique Mangubat [a] and Shane Gilroy [a]

[a]Atlantic Technological University, Sligo, Ash Lane, Co. F91 YW50, Ireland

## 1. Introduction

Cameras have become an essential component in most Autonomous Driving (AD) perception subsystems, offering high reliability in object classification, traffic sign and light recognition, and lane detection. Modern camera technology enables high-resolution imaging and rapid data acquisition, making these systems feasible for autonomous navigation [1]. Additionally, stereo cameras, combined with advanced computer vision techniques, allow for precise distance measurements, enhancing spatial awareness. Furthermore, by analyzing reference points across consecutive image frames, cameras can estimate the velocity of moving vehicles, aiding in motion prediction [2].

Despite the advantages, vision-based systems in AD face certain obstacles, particularly in challenging environmental conditions such as occlusion, low light and adverse weather. These factors can affect the accuracy of data interpretation when running computer vision algorithms, leading to potential misinterpretations that can contribute to road accidents. As image data plays a crucial role in autonomous vehicle decision-making, ensuring accurate information processing is essential. Nevertheless, cameras, similar to other sensors, have intrinsic limitations that need to be overcome to enhance the overall reliability of the system.

*1.1. Occlusion Challenges*

Efficient transportation systems rely on precise real-time target detection to facilitate effective decision-making and control signal generation. However, real-world environments present significant challenges, particularly in recognizing occluded objects. This issue is especially critical in autonomous driving, where Advanced Driver Assistance Systems (ADAS) often operate with limited computational resources. For example, as illustrated in Figure 1, an ego vehicle traveling along a roadway encounters a stationary truck on its right. The truck obstructs the view of the area behind it, concealing a cyclist attempting to cross the road. As a result, neither the vehicle driver nor the cyclist is aware of each other's presence, increasing the risk of a potential accident [3].

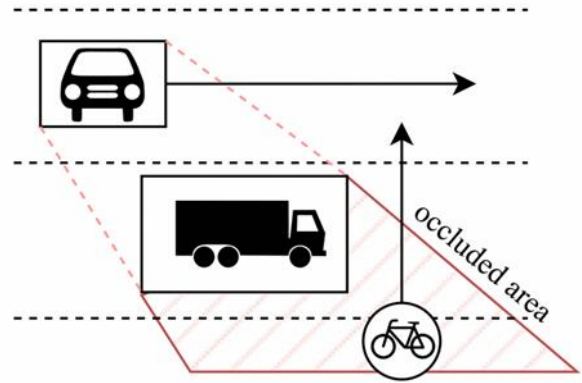

**Figure 1: Occluded bicycle scenario with an ego vehicle [3]**

A number of cyclist and bicycle detection benchmarks provide annotation for the occurrence and severity of bicycle occlusion in order to categorize algorithm performance for occluded bicycle detection such as described in [1, 2, 8, 9], however occlusion level is typically subjectively categorized into a small number of levels such as "low", "partial" and "heavy". This research proposed a novel objective method for bicycle occlusion level classification to facilitate more precise reporting of detection algorithm performance for partially occluded cyclists. The proposed methodology consists of a custom parts-based bicycle detection algorithm and an objective, repeatable method of bicycle occlusion level annotation.

## 2. Related Work

The extent to which an object is occluded, or its visibility is reduced, can be used to assess the reliability of a vision-based driver assistance systems. Previous research has focused on enhancing VRU (Vulnerable Road User) detection by improving visibility [1, 2, 4, 5, 6, 7] and categorizing different levels of VRU occlusion [1, 2, 8, 9, 10]. Ruan *et al* [11] reviewed various detection methods for occluded objects in complex driving scenarios, highlighting the similarities and differences among several occluded object detection techniques using the same sensors. The authors categorized the study of occluded object detection into three areas: occluded vehicles, pedestrians, and traffic signs. Pechinger *et al* [4] investigated cyclist safety in urban areas by applying occlusion to determine the maximum range of autonomous driving sensor systems. The study integrated microscopic traffic simulation with sub-microscopic Hardware in the Loop (HIL) simulation on the ego vehicle. The authors introduced parked road users and occluded cyclists at intersections, enabling the setup to execute an automated driving planning and control system on the vehicle computer in real time. Each scenario was assessed based on the severity of impact, categorizing situations as safe or dangerous, and considering the time to collision.

*2.1. Cyclist Datasets*

Masalov *et al* [7] introduced a publicly accessible Specialized Cyclist Dataset, dedicated exclusively to cyclist detection. This dataset was captured using a monocular RGB camera across various scenarios encountered by cyclists on roads from autumn to winter (including snowy conditions), enabling researchers to conduct thorough tests under diverse conditions. The dataset comprises 62,297 images featuring 18,200 cyclist instances and 30 different cyclists. Additionally, a specialized cyclist jersey with a diamond pattern designed to test detection accuracy against street clothes was utilized. The dataset was compared to the KITTI dataset in terms of labelling format and resolution, utilizing commonly used detection methods such as YOLO and SqueezeDet [7].

García-Venegas [2] conducted a study comparing the performance of leading deep learning-based object detection algorithms, including SSD, Faster R-CNN, and R-FCN, along with feature extractors such as InceptionV2, ResNet50, ResNet101, and Mobilenet V2, using manually annotated cyclist dataset. To determine the cyclist's direction and predict its intentions, the work proposed a multi-class detection system with eight classes based on orientation. The dataset was named "CIMAT-Cyclist," which comprises 20,229 cyclist instances across 11,103 images, labelled according to the cyclist's orientation. To enhance detection performance, a Kalman filter for tracking and the Kuhn–Munkres algorithm for multi-target association was used. Additionally, the vulnerability of cyclists was assessed in the field of view by considering the proximity and predicted intentions based on the heading angle, assigning a risk level to each cyclist [2].

Li *et al* [9] conducted an extensive experimental study on cyclist detection, evaluating current object detection methods such as Aggregated Channel Features, Deformable Part Models, and Region-based Convolutional Neural Networks. The study introduced a novel approach called Stereo-Proposal based Fast R-CNN (SP-FRCN), which utilizes stereo proposals within the Fast R-CNN (FRCN) framework for cyclist detection. The experiments were carried out on a dataset comprising 22,161 annotated cyclist instances across over 30,000 images, captured from a moving vehicle in Beijing's urban traffic. The results showed that all three method families achieved top performance with an average precision of around 0.89 in easy cases, but performance declined as difficulty increased. The dataset, named "Tsinghua-Daimler Cyclist Benchmark," included detailed annotations, stereo images, and evaluation scripts [9].

Another related study compared the performance of major deep learning-based object detection algorithms, including SSD, Faster R-CNN, and R-FCN, along with feature extractors including InceptionV2, ResNet50, ResNet101, and Mobilenet V2. To identify the cyclist's direction and predict the intentions, a multi-class detection system with eight orientation-based classes was proposed. A new dataset called "CIMAT-Cyclist," which contains 20,229 cyclist instances across 11,103 images, labelled according to the cyclist's orientation was introduced. To enhance detection performance, a Kalman filter for tracking was implemented, combined with the Kuhn–Munkres algorithm for multi-target association. The vulnerability of cyclists was assessed for each instance in the field of view, considering the proximity and predicted intentions based on the heading angle, and a risk level was assigned to each cyclist. Experimental results validated the proposed strategy in real-world scenarios [2].

In 2021, Zernetsch et al [12] presented an article on image sequence-based cyclist action recognition using multi-stream 3D convolution. Image sequences, optical flow sequences and past positions using multi-stream 3D ConvNet were used to detect the basic movements of cyclists in the real world such as: move, stop, wait and left or right turn. The work presented was intended to resolve the occlusions of cyclists by other traffic participants or road structures. The authors mounted a wide-angle stereo camera at an intersection which is heavily frequented. To be able to validate and train the new algorithm, the authors created a large dataset consisting of 1,639 video sequences with cyclists in real world traffic. In the end, 1.1 million samples were obtained. The object detection on the generated image sequences using a pre-trained object detection network trained from COCO dataset. Next, the optical flow sequence was created from the image sequence using pretrained OF network. And the trajectories were generated by tracking the head position of the cyclist in each image. The occlusions of the cyclist in the images were handled by tracking the position using Kalman Filter with constant velocity model. Abur et al [13] focused on enhancing bicycle visibility and conspicuity via increasing detection distances from. In this study, the authors investigated how to enhance visibility without necessitating any active behavior from the cyclist by analyzing the 6 rear-end components of the bicycle under different lighting conditions from manually taken images. The study used Adrian's Visibility Model to measure the visibility level of the target objects. The images were all manually annotated and evaluated by 30 respondents [13].

Other research such as [8] used bicycle detection for urban cycling infrastructure planning. In this study, a method of analyzing the demand for bicycle parking facilities using object detection on social media images was utilized. The first stage in the methodology was to use a pre-trained object detection algorithm using state-of-the-art COCO dataset to detect the bicycles on the regional subset of YFCC100m dataset collection from Flickr. Some 30, 922 images were manually annotated for stationary, parked or moving bicycles. The second stage involved using pedestrian detection to differentiate between moving and stationary bicycles [8].

Nardi et al [14] proposed a new cyclist dataset called Open Images (OI) dataset which was built from the pre-selected images of Open Images set. These images have been processed using the proposed algorithm for semiautomatic generation of cyclist annotation with the help of people and bicycle detectors. From the Open Images manually selected subset, the cyclist images were detected using YOLO object detection which was pre-trained in COCO. Thereafter, the algorithm would automatically create the annotations based on the intersection of people and bicycles. The new datasets were put to test against other datasets such as MIO-TCD, Tsinghua-Daimler and Specialized Cyclist Detection dataset using YOLO [14].

**Table 1. Cyclist Level Occlusion Summary**

| Dataset | Level of Occlusion | | | Annotation | Detection Method |
|---|---|---|---|---|---|
| | Low/No | Partial | Heavy | | |
| Tsinghua-Daimler [9] | <10% | 10-40% | 40-80% | Manual annotation with ground truth based on bounding box aspect ratios. | ACF, DPM, SP-FRCN and FRCN |
| CyDet [6] | 0% | 25% | 50% | Manually annotated ground truth. | Yolo and SqueezeDet |
| KITTI [6] | Easy | Moderate | Hard | Manually annotated ground truth. | |

A summary of occlusion level categories from different cyclist datasets is shown in Table 1. Current datasets use subjective annotation methods based on human annotation to categorize cyclists into a small number of occlusion levels.

*2.2. Parts-based Bicycle Detection*

Similar to pedestrian parts-based detection, vision-based parts detection can also be applied to bicycles [15-16]. Cho et al [15] introduced a vision-based framework for detecting and tracking bicycles, considering the changing appearance from different viewpoints while treating the rider as a non-rigid object. The authors defined a mixture of multiple viewpoints and trained a Support Vector Machine to detect bicycles in various scenarios. To enhance detection, a three-component bicycle model using Felzenszwalb's deformable part-based model and geometric concepts for each component was constructed. Finally, the cyclist was tracked using subsequent video frames with an extended Kalman filter (EKF) algorithm to estimate the position and velocity of the bicycles based on vehicle coordinates.

According to a review on the design development of bicycles [17], for bicycle safety, the typical dimensions of the bicycle include handlebar height from 0.75 to 1.10 m, handlebar width of 0.61 m, bicycle length from 1.5 to 1.8 m and tires width from 20 mm to 60 mm with contact surface of 3 mm wide. Meanwhile, the approximate dimensions of a bicycle from *Union Cycliste Internationale* (UCI) [18] are shown below in Figure 2.

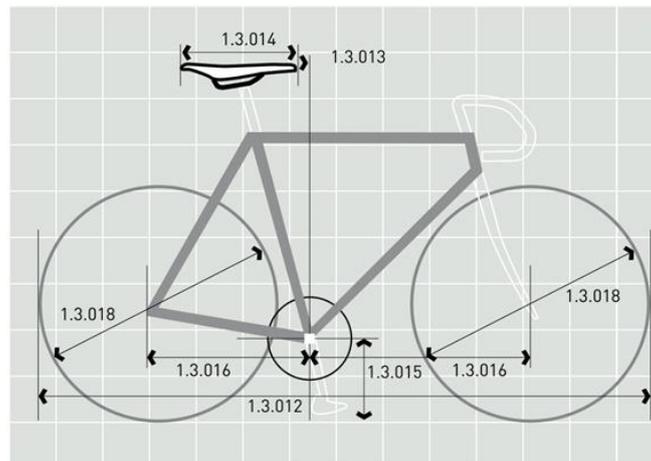

**Figure 2: Approximate bicycle dimensions from UCI [18]**

Based on the observation that a bicycle primarily consists of two elliptical wheels and a frame made up of two triangles, a paper from [16] proposed a bicycle detector for side-view images. In contrast to other classifiers that rely on feature-based analysis, the geometric constraints between the triangles and ellipses allowed for faster computation, as the algorithm required only a single image without a training process. The study claimed that the new bicycle model and algorithm offered practical and performance advantages based on experimental results. The methodology involved detecting ellipses to identify the wheels in the image. An algorithm was developed to classify arcs into four types and evaluate combinations of three arcs of different types to form an ellipse. The arcs were then classified by slope and convexity. Subsequently, the frame of the target bicycle was generated from the extracted triangles within the search region. The width and height of the search region were determined by the distance between the two detected ellipses and the length of the major axis of the detected ellipses, respectively. To verify if the two detected ellipses were bicycle wheels, the relationship between the detected frame with two triangles and the two detected wheels was established.

Chen *et al* [19] published a paper on a vision-based nighttime detection method for bicycles and motorcycles, utilizing a camera and near-infrared lighting mounted on an automobile. The authors explained that foreground objects reflect near-infrared lighting in nighttime environments, while the distant background does not. However, components of bicycles and motorcycles sometimes absorb most of the infrared lighting, making them difficult to recognize. To address this issue, a part-based detection method that combines features specific to bicycles and motorcycles was developed. Information on the geometric relationships among all parts and the object centroid was learned offline. Due to the high computational load, the Adaboost algorithm was employed to select effective parts with better geometric information for detection. The new approach described bicycles and motorcycles using part-based features, which included appearance-based and edge-based features. For appearance-based features, patches were defined as fixed-ratio rectangular regions containing keypoints in positive training samples, encoded with HOG features, and trained using linear SVM classifiers, as exemplified in Figure 3. Edge-based features were inspired by the Boundary Fragment Model (BFM), which encodes segments of linking edges on the object. The part-based construction involved three steps: part extraction, part clustering, and part selection.

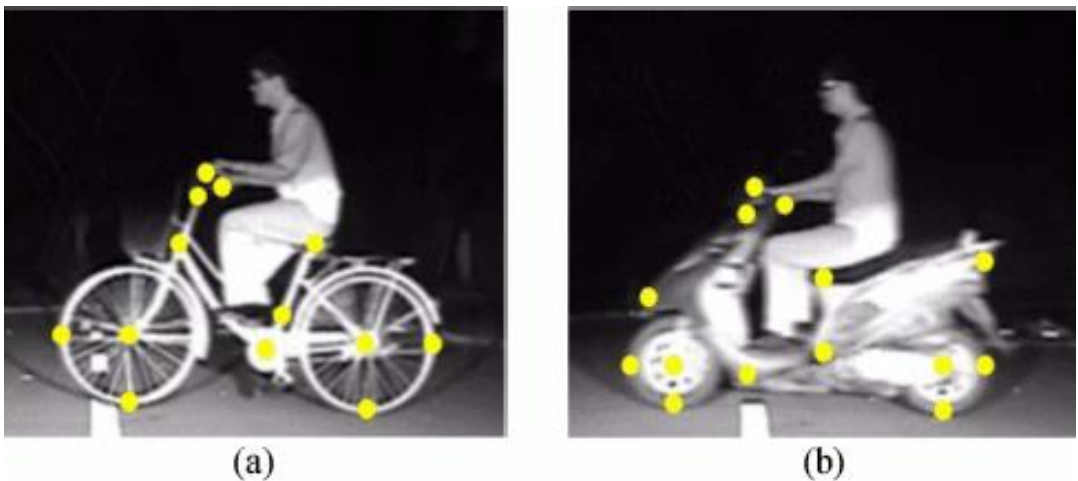

**Figure 3: Keypoint detection on Bicycle (A) and Motorcycle (B) at night [19]**

Numerous research has been done with regards to object detection, particularly vulnerable road users (VRUs) such as pedestrians and cyclists. However, these studies use subjective interpretation of how to gauge the level of object visibility and occlusion. There is no standard procedure on how to measure and compare visibility or occlusion levels used in different object detection algorithms, even less for bicycle parts detection. In order to compare different object detection methods, an objective method or process that calculates and classifies a bicycle's level of visibility or occlusion in image detection is required. This research addresses this significant knowledge gap.

## 3. Methodology

Inspired by the work of [23-25], this study aims to improve road safety by introducing a novel objective method of bicycle occlusion level classification that can be used to quantify the visibility of bicycles and cyclists in images. To achieve this, the proposed methodology uses a novel parts-based detection model and analyzes occlusion and visibility level using geometric data. The research follows a quantitative approach, focusing on numerical data from detection algorithms. An experimental design is used to develop the detection model and benchmark, which is tested across different scenarios. Data is sourced from existing types of mobility study [20] and the Tsinghua-Daimler cyclist dataset [21], with images allocated into training, validation, and test sets. A quantitative analysis evaluates the benchmark's performance against prior research, ensuring its reliability. Ethical considerations confirm that datasets are used in accordance with academic licensing terms.

*3.1. Deformable Parts-Based Model*

A bicycle detection dataset is curated from existing works and re-annotated using the Roboflow software platform [26]. New annotation labels are provided to identify different bicycle parts: wheel, frame, and handlebar, by drawing polygons over the parts and labelling them accordingly. The annotated images are saved in a new dataset and divided into training, validation, and test sets. Additional images are generated by augmenting the originals, altering factors such as grayscale, brightness (from -20% to 20%), blur (up to 2.5px), and noise (up to 1.37% of pixels). The new dataset is pre-processed to fit a 640x640 size before undergoing training. The resulting model is then used as a detection model within a workflow. The workflow reads an image, applies the detection model and outputs the inference data such as the predicted bicycle parts, bounding box dimension and confidence level in a JSON format file. This output file is used in a python-based algorithm which extracts the predicted bicycle parts [27]. The python code assigns the visibility value for each bicycle parts detected according to the percentage distribution in Figure 4.

The visibility allocation is based on the standard bicycle parts specification available in the market. The standard 26-inch wheel has an approximate area of 3400cm$^2$, a standard frame 1454cm2 and a typical handlebar has approximately 110cm$^2$ [18]. The total visible bicycle surface area is calculated by dividing each part with the total area of 8364cm$^2$.

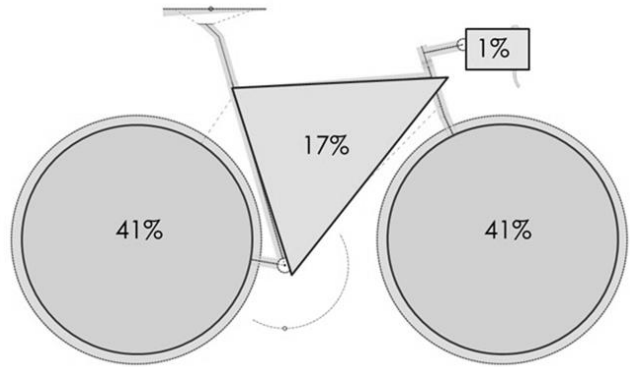

**Figure 4: Bicycle Surface Area**

*3.2. Occlusion Level Implementation*

An overview of the occlusion level classification method is shown in Figure 5. The algorithm searches for the predicted bicycle parts from the JSON output file and assigns the visibility value based on the visibility bicycle surface area, Figure 4. Meanwhile, for the two wheels, the python code compares the x and y dimension of the bounding box to check if the wheel is at an angle or partially visible before assigning the visibility level. The visibility level of each individual part is then combined to get the total bicycle visibility. Lastly, the bicycle occlusion level is determined by subtracting the total bicycle visibility from the total surface area as shown in Figure 4.

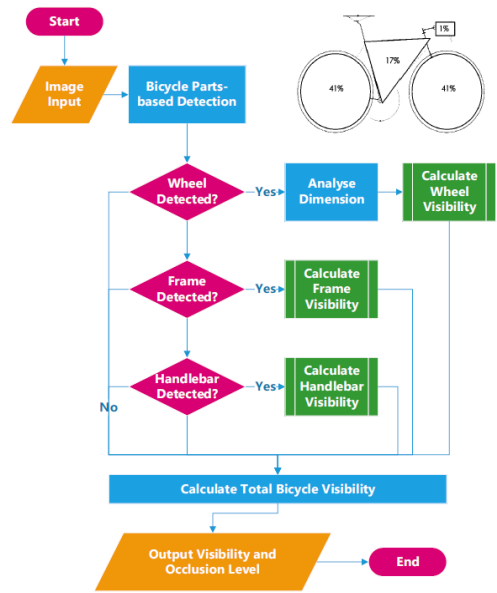

**Figure 5: Bicycle Occlusion Level Flowchart**

Images of bicycles with varying degrees of occlusion are gathered to form a test dataset. These images are processed through the detection workflow and the resulting JSON files are collected. The detection output is then processed by the algorithm. Finally, the visibility of bicycle parts and occlusion levels are systematically tabulated and examined for further insight.

## 4. Results and Discussion

The primary objective of these experiments is to assess the performance of the proposed parts-based detection model and quantify bicycle occlusion level. Specific metrics such as detection accuracy, occlusion measurement, and model confidence serve as the focal points of the analysis.

*4.1. Model Performance*

The combined re-annotated images from the Types of Active Mobility [20] and Tsinghua-Daimler [21] datasets were trained using instance segmentation detection. The trained model achieved a mean Average Precision (mAP@50) of 59.8% with a 50% confidence threshold, a precision of 83.4%, and a recall of 48.0%. In contrast, training the Types of Active Mobility dataset alone resulted in a mAP@50 of 89.5%, precision of 91.7%, and recall of 85.2%. Table 2 summarizes the mAP by class. Both validation and test sets showed that the handlebar had the lowest accuracy, contributing to the lower overall mAP value.

Table 2. Average Precision by Class

| mAP@50 | Combined Dataset | | Active Mobility Dataset | |
|---|---|---|---|---|
| | Validation Set | Test Set | Validation Set | Test Set |
| Frame | 68% | 67% | 90% | 92% |
| Handlebar | 34% | 32% | 82% | 73% |
| Wheel | 78% | 78% | 96% | 96% |
| Mean | 60% | 59% | 90% | 87% |

Meanwhile, the confusion matrix displays the false negative and false positive values by comparing the actual ground truth with the model predictions at mAP@50 as seen in Tables 3 and 4. The diagonal of the matrix represents instances where the model correctly predicted a class in the test or validation image set. Other predictions are either model errors or labelling errors, resulting in false positives or false negatives.

Table 3. Combined Dataset Model Training Confusion Matrix

| Ground Truth | Prediction | | | |
|---|---|---|---|---|
| | frame | handlebar | wheel | False Negative |
| frame | 84 | 0 | 0 | 171 |
| handlebar | 0 | 47 | 0 | 208 |
| wheel | 2 | 0 | 403 | 523 |
| False Positive | 12 | 9 | 6 | 0 |

Table 4. Active Mobility Dataset Model Training Confusion Matrix

| Ground Truth | Prediction | | | |
|---|---|---|---|---|
| | frame | handlebar | wheel | False Negative |
| frame | 50 | 0 | 0 | 9 |
| handlebar | 0 | 36 | 0 | 14 |
| wheel | 0 | 0 | 106 | 12 |
| False Positive | 3 | 4 | 4 | 0 |

After training, the models are used to test images to extract the predicted classes, confidence level and the bounding box dimensions. Both the trained segmentation or parts-based model from combined dataset and active mobility dataset show the same class predictions but slightly different confidence values as seen in Figure 6. Additionally, the outlined dimensions for the bounding boxes are also slightly different. Although the first model seems to give higher confidence in the predicted classes, the later model detects more bicycle parts when tested on multiple bicycles in one image frame.

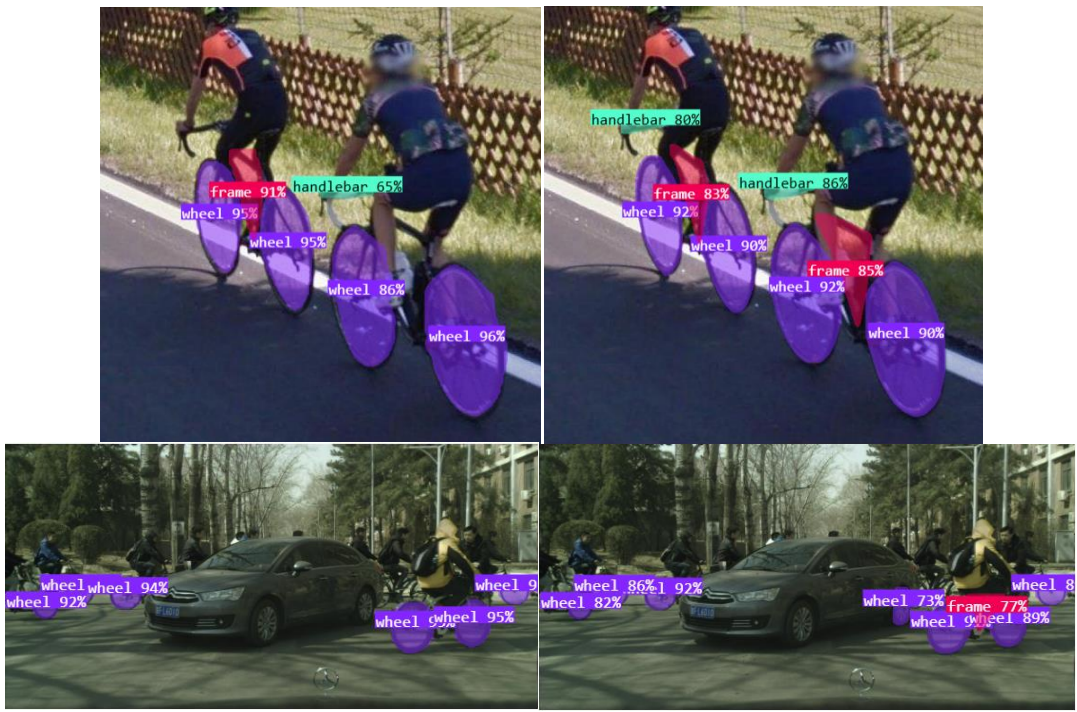

**Figure 6: Multiple bicycle predictions for Combined Dataset (Left) and Active Mobility Dataset (Right) with Level of Confidence.**

*4.2. Occlusion Level Application*

After the detection models generate predictions, the output is converted into bicycle part data, which is then interpreted by the Python code. Each bicycle part is analyzed to determine its visibility value. This section focuses on the results of applying geometric analysis to the detected bicycle parts. Figure 7 compiles selected images showing various states of bicycle parts, such as angled, hidden, or occluded. Table 5 summarizes the visibility and occlusion levels for each image.

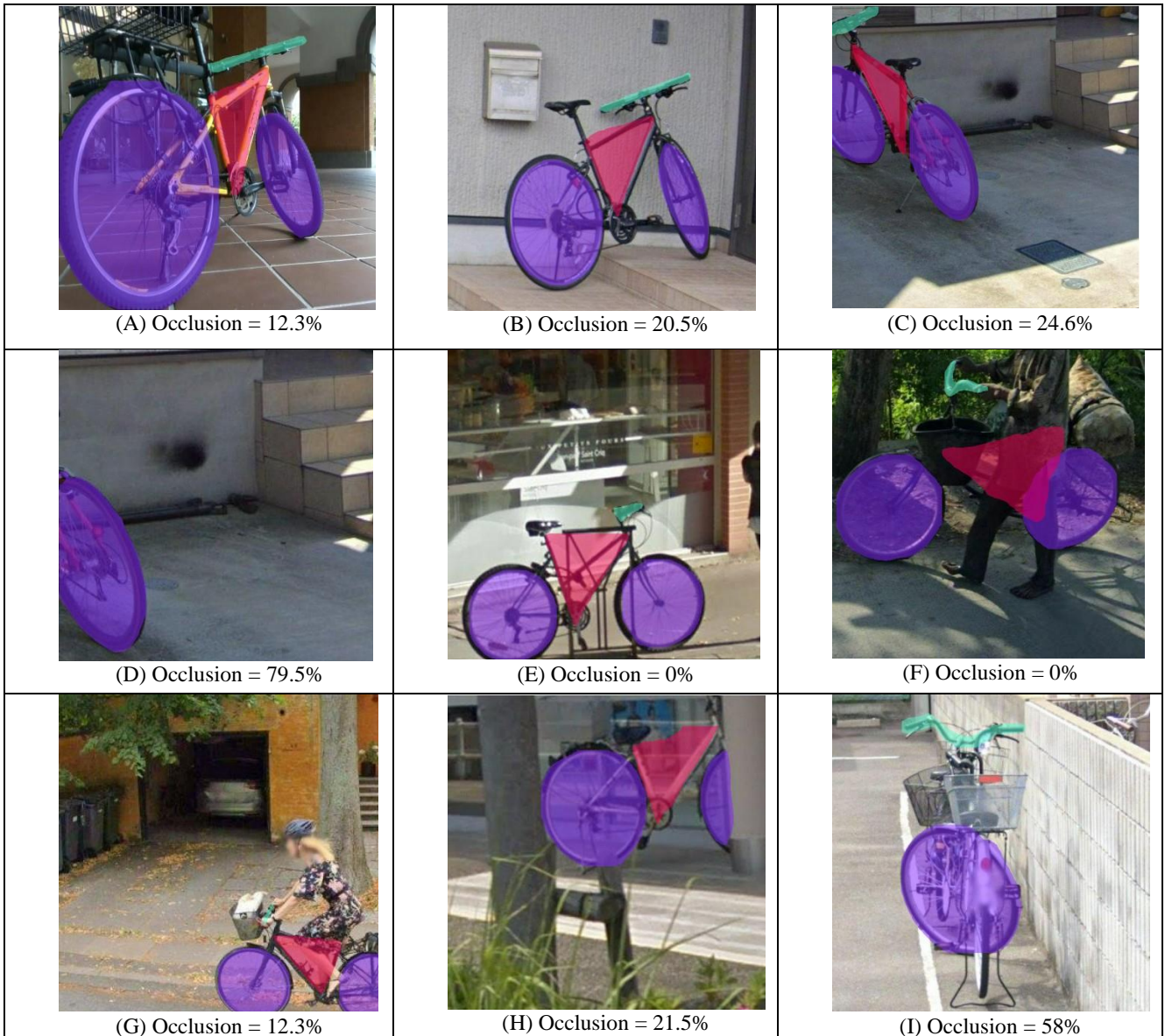

**Figure 7: Images of bicycle parts-based detection in different occlusion scenarios**

**Table 5. Summary of Bicycle Visibility and Occlusion Values for different scenarios.**

| Scenario | Wheel (%) | Frame (%) | Handlebar (%) | Bicycle Visibility (%) | Bicycle Occlusion (%) |
|---|---|---|---|---|---|
| A | 69.7 | 17 | 1 | 87.7 | 12.3 |
| B | 61.5 | 17 | 1 | 79.5 | 20.5 |
| C | 57.4 | 17 | 1 | 75.4 | 24.6 |
| D | 20.5 | 0 | 0 | 20.5 | 79.5 |
| E | 82.0 | 17 | 1 | 100.0 | 0.0 |
| F | 82.0 | 17 | 1 | 100.0 | 0.0 |
| G | 69.7 | 17 | 1 | 87.7 | 12.3 |
| H | 61.5 | 17 | 0 | 78.5 | 21.5 |
| I | 41.0 | 0 | 1 | 42.0 | 58.0 |

Wheel visibility varies significantly across scenarios, indicating a strong dependence on specific conditions set by the author in the x and y dimensions of the detected bounding box. Generally, wheels and handlebars are more visible compared to other parts. Overall bicycle visibility ranges from 20.5% to 100%, with an average of 74.59%. Occlusion levels also vary widely, from 0.0% to 79.5%, reflecting different degrees of obstruction in various scenarios.

The experimental data is analyzed to uncover further insight into the behavior of the detection model under varying conditions. Special emphasis is placed on interpreting how these results inform our understanding of bicycle occlusion in real-world scenarios. For the model qualitative results, the combined datasets show poorer results on both misclassifications of false positives and false negatives as well as edge cases with multiple bicycles compared to the active mobility datasets. However, both models still give significantly high precision and confidence values demonstrating a level of robustness during detection. The F1-score and mAP values could be enhanced by varying levels of confidence threshold. One limitation of the study is the reliance on a relatively small dataset, which may not capture the full variability of real-world scenarios. Additionally, the model's performance on low-resolution images suggests a need for further refinement. In summary, the developed parts-based model demonstrates strong performance in image classification tasks. The new benchmark is able to detect a bicycle part and identify its level of occlusion. The findings highlight the potential for practical applications and provide a foundation for future research in this area.

## 5. Conclusion

This research proposes a novel method of evaluating and classifying bicycle occlusion levels, with the aim of enhancing cyclist safety by improving detection accuracy through a parts-based detection model. Analysis demonstrates that the proposed model not only quantifies occlusion levels but also reliably distinguishes between various bicycle sematic parts under diverse conditions. These results underscore the potential of computer vision techniques in revolutionizing road safety standards, paving the way for more informed infrastructure design and safety protocols. While the results are promising, the study is confined by the available dataset size and controlled testing conditions, which may not fully capture the variability of real-world scenarios. The findings indicate that occlusion levels can vary significantly based on factors such as part visibility, shape, angle, and image quality. To enhance model performance in future research, it is recommended to refine the annotation process. Future work will expand on this benchmark by incorporating larger, more varied datasets and refining the model to better handle dynamic environmental conditions using diverse images and improving the model's robustness to low-quality inputs. Exploring different architectures and training techniques could also yield further improvements. Integration of the proposed methodology with the objective pedestrian occlusion level classification method outlined by Gilroy *et al* [24] will provide a comprehensive cyclist detection classification method for partially occluded cyclists. In conclusion, the development of this benchmark marks a significant step towards more accurate and effective bicycle detection systems, ultimately contributing to safer roads and more resilient urban mobility frameworks.